\documentclass[11pt]{article}

\usepackage{colacl}
\usepackage{times}
\usepackage{epsf}

\newcommand{\FOR}{\textbf{for}}
\newcommand{\TO}{\textbf{to}}
\newcommand{\IF}{\textbf{if}}
\newcommand{\WHILE}{\textbf{while}}

\newcommand{\MAXREAL}{\mathrm{MAXREAL}}
\newcommand{\len}{\mathit{len}}
\newcommand{\lenst}{\mathrm{len}}
\newcommand{\simst}{\mathit{sim}}
\newcommand{\cost}{\mathit{cost}}
\newcommand{\locc}{\mathit{loc\_c}}
\newcommand{\link}{\mathit{link}}
\newcommand{\parst}{\mathit{par}}

\newcommand{\cum}{\mathrm{sum}}
\newcommand{\prev}{\mathrm{prev}}
\newcommand{\avgst}{\mathrm{avg}}
\newcommand{\minst}{\mathrm{min}}
\newcommand{\maxst}{\mathrm{max}}

\title{Optimal Multi-Paragraph Text Segmentation by Dynamic Programming}

\author{Oskari Heinonen\\[1ex]
University of Helsinki, Department of Computer Science\\
P.O.\,Box 26 (Teollisuuskatu 23), FIN--00014 University of Helsinki,
Finland\\[1ex]
\emph{Oskari.Heinonen@cs.Helsinki.FI}}

\begin{document}

\maketitle


\begin{abstract}
There exist several methods of calculating a similarity curve,
or a sequence of similarity values, representing the lexical cohesion of
successive text constituents, e.g., paragraphs.  Methods for deciding
the locations of fragment boundaries are, however, scarce.  We propose a
fragmentation method based on dynamic programming.  The method is
theoretically sound and guaranteed to provide an optimal splitting on
the basis of a similarity curve, a preferred fragment length, and a cost
function defined.  The method is especially useful when control on
fragment size is of importance.
\end{abstract}

\section{Introduction}
\label{s:introduction}

Electronic full-text documents and digital libraries make the
utilization of texts much more effective than before; yet, they pose new
problems and requirements.  For example, document retrieval based on
string searches typically returns either the whole document or just the
occurrences of the searched words.  What the user often is after,
however, is \emph{microdocument}: a part of the document that contains
the occurrences and is reasonably self-contained. 

Microdocuments can be created by utilizing lexical cohesion (term
repetition and semantic relations) present in the text.  There exist
several methods of calculating a \emph{similarity curve}, or a sequence
of similarity values, representing the lexical cohesion of successive
constituents (such as paragraphs) of text (see, e.g.,
\cite{Hearst:ACL94,Hearst:CL97,Kozima:ACL93,MorrisHirst:CL91,%
Yaari:RANLP97,Youmans:L91}). 
Methods for deciding the locations of fragment boundaries are, however,
not that common, and those that exist are often rather heuristic in nature. 

To evaluate our fragmentation method, to be explained in
Section~\ref{s:fragmentation}, we calculate the paragraph similarities
as follows.  We employ stemming, remove stopwords, and count the
frequencies of the remaining words, i.e., terms.  Then we take a
predefined number, e.g., 50, of the most frequent terms to represent the
paragraph, and count the similarity using the cosine coefficient (see,
e.g., \cite{Salton:89}).
Furthermore, we have applied a sliding window method: instead of just
one paragraph, several paragraphs on both sides of each paragraph
boundary are considered.  The paragraph vectors are weighted based on
their distance from the boundary in question with immediate paragraphs
having the highest weight.  The benefit of using a larger window is that
we can smooth the effect of short paragraphs and such, perhaps
example-type, paragraphs that interrupt a chain of coherent paragraphs. 

\section{Fragmentation by Dynamic Programming}
\label{s:fragmentation}

\begin{figure}[t]
\begin{small}
\begin{center}
\begin{minipage}{0pt}
\begin{tabbing}
xxx \= xxx \= xxx \= xxx \= xxx \= xxx \= xxx \= xxx \= xxx \= xxx \= \kill
fragmentation($n$, $p$, $h$, $\len[1 .. n]$, $\simst[1 .. n - 1]$)\\
/* $n$ no.\ of pars, $p$ preferred frag length, $h$ scaling */\\
/* $\len[1 .. n]$ par lengths, $\simst[1 .. n - 1]$ similarities */\\
\{\\
\> $\simst[0] := 0.0$; $\cost[0] := 0.0$; $B := \emptyset$;\\
\> \FOR\ $\parst := 1$ \TO\ $n$ \{\\
\> \> $\len_{\cum} := 0$; /* cumulative fragment length */\\
\> \> $c_{\minst} := \MAXREAL$;\\
\> \> \FOR\ $i := \parst$ \TO\ $1$ \{\\
\> \> \> $\len_{\cum} := \len_{\cum} + \len[i]$;\\
\> \> \> $c := c_{\lenst}$($\len_{\cum}$, $p$, $h$);\\
\> \> \> \IF\ $c > c_{\minst}$ \{ /* optimization */\\
\> \> \> \> exit the innermost for loop;\\
\> \> \> \}\\
\> \> \> $c := c + \cost[i - 1] + \simst[i - 1]$;\\
\> \> \> \IF\ $c < c_{\minst}$ \{\\
\> \> \> \> $c_{\minst} := c$; $\locc_{\minst} := i - 1$;\\
\> \> \> \}\\
\> \> \}\\
\> \> $\cost[\parst] := c_{\minst}$; $\link_{\prev}[\parst] :=
\locc_{\minst}$;\\
\> \}\\
\> $j := n$;\\
\> \WHILE\ $\link_{\prev}[j] > 0$ \{\\
\> \> $B :=  B \cup \link_{\prev}[j]$; $j := \link_{\prev}[j]$;\\
\> \}\\
\> return($B$); /* set of chosen fragment boundaries */\\
\}
\end{tabbing}
\end{minipage}
\end{center}
\end{small}
\caption{The dynamic programming algorithm for fragment boundary detection.}
\label{f:splitting}
\end{figure}

Fragmentation is a problem of choosing the paragraph boundaries that
make the best fragment boundaries.  The local minima of the similarity
curve are the points of low lexical cohesion and thus the natural
candidates.  To get reasonably-sized microdocuments, the similarity
information alone is not enough; also the lengths of the created
fragments have to be considered.  In this section, we describe an
approach that performs the fragmentation by using both the
similarities and the length information in a robust manner.  The
method is based on a programming paradigm called dynamic programming
(see, e.g., \cite{CormenEtAl:90}).  Dynamic programming as a method
guarantees the optimality of the result with respect to the input and
the parameters.

The idea of the fragmentation algorithm is as follows (see also
Fig.~\ref{f:splitting}).  We start from the first boundary and calculate
a cost for it as if the first paragraph was a single fragment.  Then we
take the second boundary and attach to it the minimum of the two
available possibilities: the cost of the first two paragraphs as if they
were a single fragment and the cost of the second paragraph as a
separate fragment.  In the following steps, the evaluation moves on by
one paragraph at each time, and all the possible locations of the
previous breakpoint are considered.  We continue this procedure till the
end of the text, and finally we can generate a list of breakpoints that
indicate the fragmentation. 

The cost at each boundary is a combination of three components: the
cost of fragment length $c_{\lenst}$, and the cost $\cost[\cdot]$ and
similarity $\simst[\cdot]$ of some previous boundary.  The cost
function $c_{\lenst}$ gives the lowest cost for the preferred fragment
length given by the user, say, e.g., 500 words.  A fragment which is
either shorter or longer gets a higher cost, i.e., is punished for its
length.  We have experimented with two families of cost functions, a
family of second degree functions (parabolas),
$$\textstyle c_{\lenst}(x, p, h) = h(\frac{1}{p^2}x^{2} - \frac{2}{p}x + 1),$$
and V-shape linear functions,
$$\textstyle c_{\lenst}(x, p, h) = |h(\frac{x}{p}-1)|,$$
where $x$ is the actual fragment length, $p$ is the preferred fragment
length given by the user, and $h$ is a scaling parameter that allows
us to adjust the weight given to fragment length.  The smaller the
value of $h$, the less weight is given to the preferred fragment
length in comparison with the similarity measure.

\section{Experiments}
\label{s:experiments}

\begin{figure}[t]
\begin{small}
\begin{center}
\epsfxsize=\columnwidth
\leavevmode
\epsffile{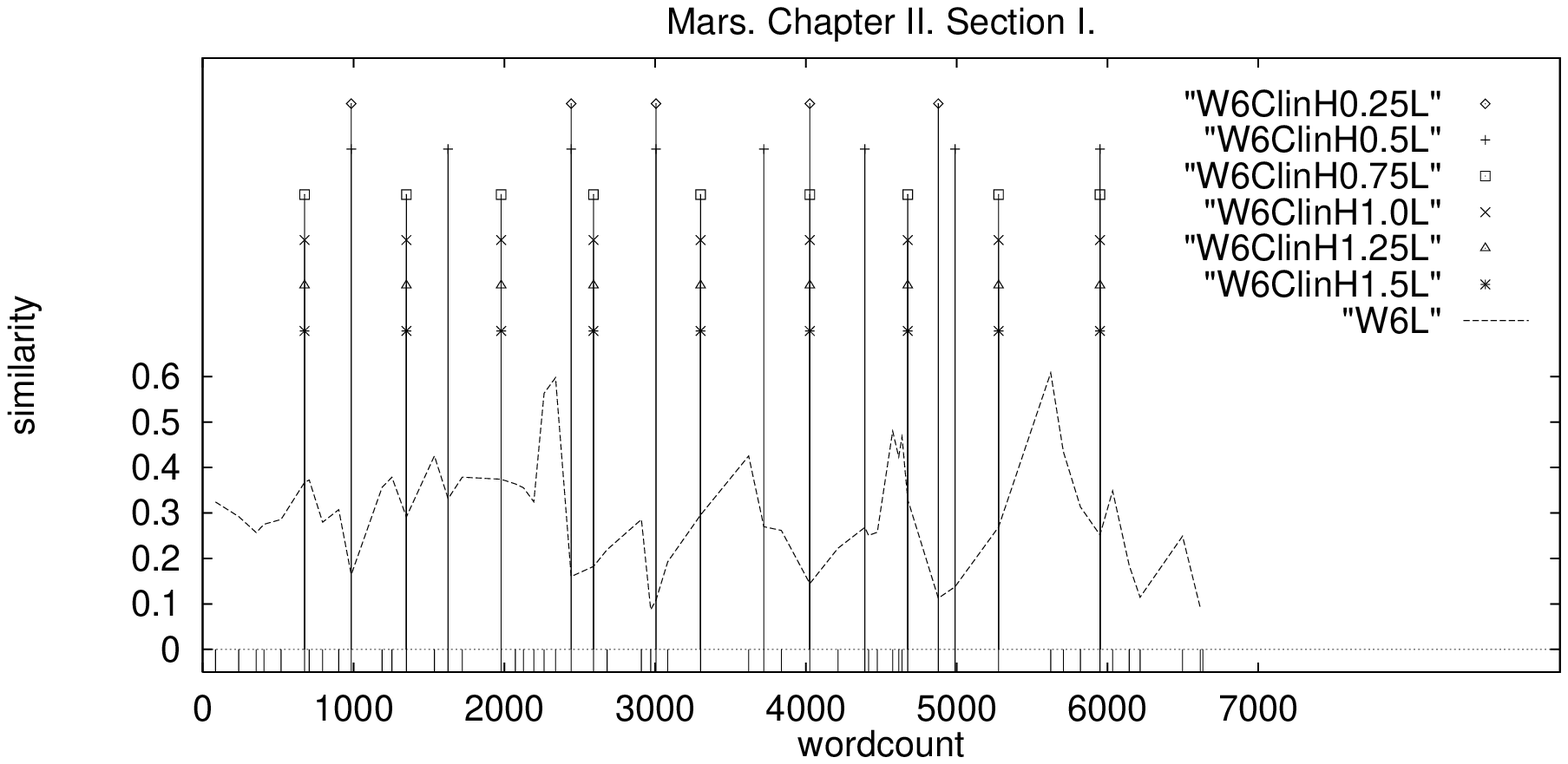}\\
(a)\\[1ex]
\epsfxsize=\columnwidth
\leavevmode
\epsffile{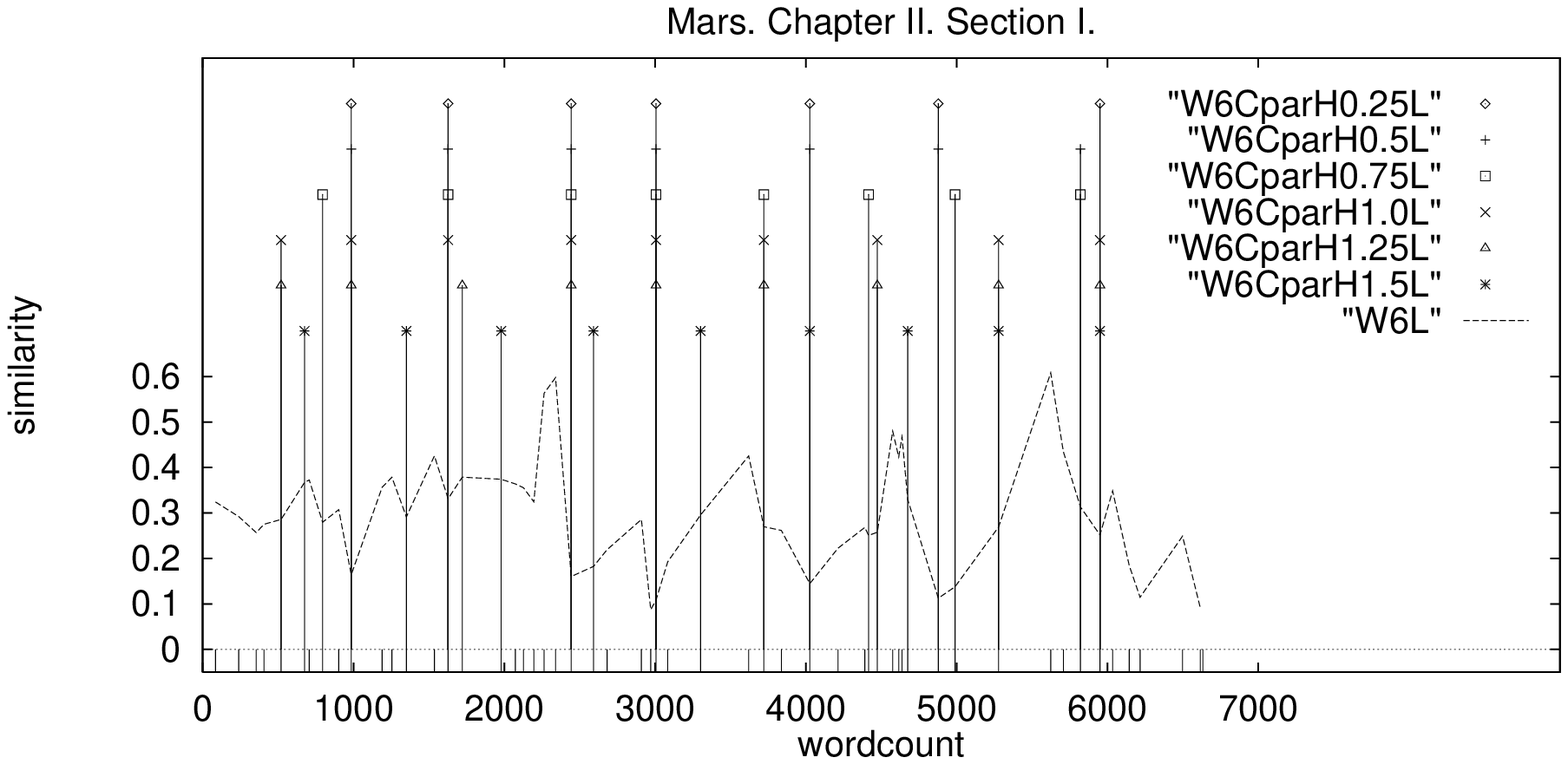}\\
(b)
\end{center}
\end{small}
\caption{Similarity curve and detected fragment boundaries
with different cost functions. 
(a)~Linear.  (b)~Parabola.  $p$ is 600 words in both (a)~\&~(b). 
``H0.25'', etc., indicates the value of~$h$. Vertical bars indicate
fragment boundaries while short bars below horizontal axis indicate paragraph
boundaries.} 
\label{f:cost2}
\end{figure}

As test data we used \emph{Mars} by Percival Lowell, 1895.
As an illustrative example, we
present the analysis of Section I.\ \emph{Evidence of it} of Chapter
II.\ \emph{Atmosphere}.  The length of the section is approximately 6600
words and it contains 55 paragraphs.
The fragments
found with different parameter settings can be seen in
Figure~\ref{f:cost2}.  One of the most interesting is the one with
parabola cost function and $h = .5$.  In this case the fragment length
adjusts nicely according to the similarity curve.  Looking at the text,
most fragments have an easily identifiable topic, like atmospheric
chemistry in fragment~7.  Fragments 2 and~3 seem to have roughly the
same topic: measuring the diameter of the planet Mars.  The fact that
they do not form a single fragment can be explained by the preferred
fragment length requirement. 

\begin{table}[t]
\begin{small}
\begin{tabular*}{\columnwidth}{l|r|rrrr}
\multicolumn{1}{c|}{cost function}&\multicolumn{1}{c|}{$h$}&%
\multicolumn{1}{c}{$l_\avgst$}&\multicolumn{1}{c}{$l_\minst$}&%
\multicolumn{1}{c}{$l_\maxst$}&\multicolumn{1}{c}{$d_\avgst$}\\\hline\hline
linear&.25&1096.1&501&3101&476.5\\
&.50&706.4&501&1328&110.5\\
&.75&635.7&515&835&60.1\\
&1.00&635.7&515&835&59.5\\
&1.25&635.7&515&835&59.5\\
&1.50&635.7&515&835&57.6\\\hline
parabola&.25&908.2&501&1236&269.4\\
&.50&691.0&319&1020&126.0\\
&.75&676.3&371&922&105.8\\
&1.00&662.2&371&866&94.2\\
&1.25&648.7&466&835&82.4\\
&1.50&635.7&473&835&69.9
\end{tabular*}
\end{small}
\caption{Variation of fragment length.  Columns: $l_\avgst$, $l_\minst$,
$l_\maxst$ average, minimum, and maximum fragment length; and $d_\avgst$
average deviation.}
\label{t:costtable}
\end{table}

Table~\ref{t:costtable} summarizes the effect of the scaling factor $h$
in relation to the fragment length variation with the two cost functions
over those 8 sections of \emph{Mars} that have a length of at least 20
paragraphs.  The average deviation $d_\avgst$ with respect to the
preferred fragment length $p$ is defined as
$d_\avgst = (\sum_{i=1}^{m} |p - l_i|)/m$
where $l_i$ is the length of fragment $i$, and $m$ is the number of
fragments.  The parametric cost function chosen affects the result a
lot.  As expected, the second degree cost function allows more
variation than the linear one but roles change with a small $h$.
Although the experiment is insufficient, we can see that in this
example a factor $h \geq 1.0$ is unsuitable with the linear cost
function (and $h = 1.5$ with the parabola) since in these cases so
much weight is given to the fragment length that fragment boundaries
can appear very close to quite strong local maxima of the similarity
curve.

\section{Conclusions}
\label{s:conclusions}

In this article, we presented a method for detecting fragment boundaries
in text.  The fragmentation method is based on dynamic programming and
is guaranteed to give an optimal solution with respect to a similarity
curve, a preferred fragment length, and a parametric fragment-length
cost function defined.  The method is independent of the similarity
calculation.  This means that any method, not necessarily based on
lexical cohesion, producing a suitable sequence of similarities can be
used prior to our fragmentation method.  For example, the \emph{lexical
cohesion profile} \cite{Kozima:ACL93} should be perfectly usable with
our fragmentation method. 

The method is especially useful when control over fragment size is
required.  This is the case in passage retrieval since windows of 1000
bytes \cite{WilkinsonZobel:TREC95} or some hundred words
\cite{Callan:SIGIR94} have been proposed as best passage sizes.
Furthermore, we believe that fragments of reasonably similar size are
beneficial in our intended purpose of document assembly.

\section*{Acknowledgements}

This work has been supported by the Finnish Technology Development
Centre (TEKES) together with industrial partners, and by a grant from
the 350th Anniversary Foundation of the University of Helsinki.  The
author thanks Helena Ahonen, Barbara Heikkinen, Mika Klemettinen, and
Juha K{\"a}rkk{\"a}inen for their contributions to the work described.



\begin{thebibliography}{}

\bibitem[\protect\citename{Callan}1994]{Callan:SIGIR94}
J.~P. Callan.
\newblock 1994.
\newblock Passage-level evidence in document retrieval.
\newblock In {\em Proc.\ SIGIR'94}, Dublin, Ireland.

\bibitem[\protect\citename{Cormen \bgroup et al.\egroup }1990]{CormenEtAl:90}
T.~H. Cormen, C.~E. Leiserson, and R.~L. Rivest.
\newblock 1990.
\newblock {\em Introduction to Algorithms}.
\newblock MIT Press, Cambridge, MA, USA.

\bibitem[\protect\citename{Hearst}1994]{Hearst:ACL94}
M.~A. Hearst.
\newblock 1994.
\newblock Multi-paragraph segmentation of expository text.
\newblock In {\em Proc.\ ACL-94}, Las Cruces, NM, USA.

\bibitem[\protect\citename{Hearst}1997]{Hearst:CL97}
M.~A. Hearst.
\newblock 1997.
\newblock Text{T}iling: Segmenting text into multi-paragraph subtopic
  passages.
\newblock {\em Computational Linguistics}, 23(1):33--64, March.

\bibitem[\protect\citename{Kozima}1993]{Kozima:ACL93}
H.~Kozima.
\newblock 1993.
\newblock Text segmentation based on similarity between words.
\newblock In {\em Proc.\ ACL-93}, Columbus, OH, USA.

\bibitem[\protect\citename{Morris and Hirst}1991]{MorrisHirst:CL91}
J.~Morris and G.~Hirst.
\newblock 1991.
\newblock Lexical cohesion computed by thesaural relation as an indicator of
  the structure of text.
\newblock {\em Computational Linguistics}, 17(1):21--48.

\bibitem[\protect\citename{Salton}1989]{Salton:89}
G.~Salton.
\newblock 1989.
\newblock {\em Automatic Text Processing: The Transformation, Analysis, and
  Retrieval of Information by Computer}.
\newblock Addison-Wesley, Reading, MA, USA.

\bibitem[\protect\citename{Wilkinson and Zobel}1995]{WilkinsonZobel:TREC95}
R.~Wilkinson and J.~Zobel.
\newblock 1995.
\newblock Comparison of fragmentation schemes for document retrieval.
\newblock In {\em Overview of TREC-3}, Gaithersburg, MD, USA.

\bibitem[\protect\citename{Yaari}1997]{Yaari:RANLP97}
Y.~Yaari.
\newblock 1997.
\newblock Segmentation of expository texts by hierarchical agglomerative
  clustering.
\newblock In {\em Proc.\ RANLP'97}, Tzigov Chark, Bulgaria.

\bibitem[\protect\citename{Youmans}1991]{Youmans:L91}
G.~Youmans.
\newblock 1991.
\newblock A new tool for discourse analysis.
\newblock {\em Language}, 67(4):763--789.

\end{thebibliography}

\begin{onecolumn}

\section*{Errata}

\noindent Table~\ref{t:costtable} is incorrect.
Table~\ref{t:costtablei} is correct.

\noindent Figure~\ref{f:cost2} is tiny.
Figure~\ref{f:cost2i} has been enlarged.

\noindent Figure~\ref{f:costpar25} here is additional.

\vfill

\begin{table}[h]
\begin{center}
\begin{tabular}{l|r|rrrr}
\multicolumn{1}{c|}{cost function}&\multicolumn{1}{c|}{$h$}&%
\multicolumn{1}{c}{$l_\avgst$}&\multicolumn{1}{c}{$l_\minst$}&%
\multicolumn{1}{c}{$l_\maxst$}&\multicolumn{1}{c}{$d_\avgst$}\\\hline\hline
linear&.25&1105.3&562&1754&518.0\\
&.50&736.9&562&985&145.8\\
&.75&663.2&603&724&63.2\\
&1.00&663.2&603&724&63.2\\
&1.25&663.2&603&724&63.2\\
&1.50&663.2&603&724&63.2\\\hline
parabola&.25&829.0&562&1072&238.5\\
&.50&829.0&562&1020&238.5\\
&.75&736.9&562&832&151.3\\
&1.00&663.2&466&817&113.8\\
&1.25&663.2&466&804&113.8\\
&1.50&663.2&603&724&63.2
\end{tabular}
\end{center}
\caption{Variation of fragment length.  Columns: $l_\avgst$, $l_\minst$,
$l_\maxst$ average, minimum, and maximum fragment length; and $d_\avgst$
average deviation.}
\label{t:costtablei}
\end{table}

\vfill

\begin{figure}[h]
\begin{center}
\epsfxsize=.75\textwidth
\leavevmode
\epsffile{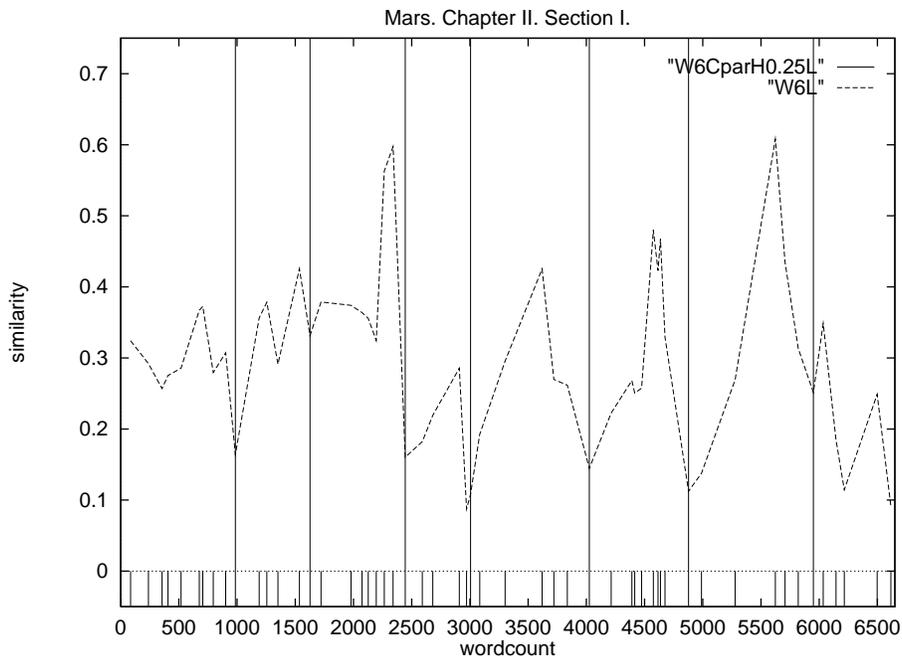}\\[1ex]
\end{center}
\caption{Similarity curve and detected fragment boundaries.  Parabola
cost function, $p$ is 600 words, and $h = .25$.  Vertical bars indicate
fragment boundaries while short bars below horizontal axis indicate
paragraph boundaries.}
\label{f:costpar25}
\end{figure}

\begin{figure}[p]
\begin{center}
\epsfxsize=\textwidth
\leavevmode
\epsffile{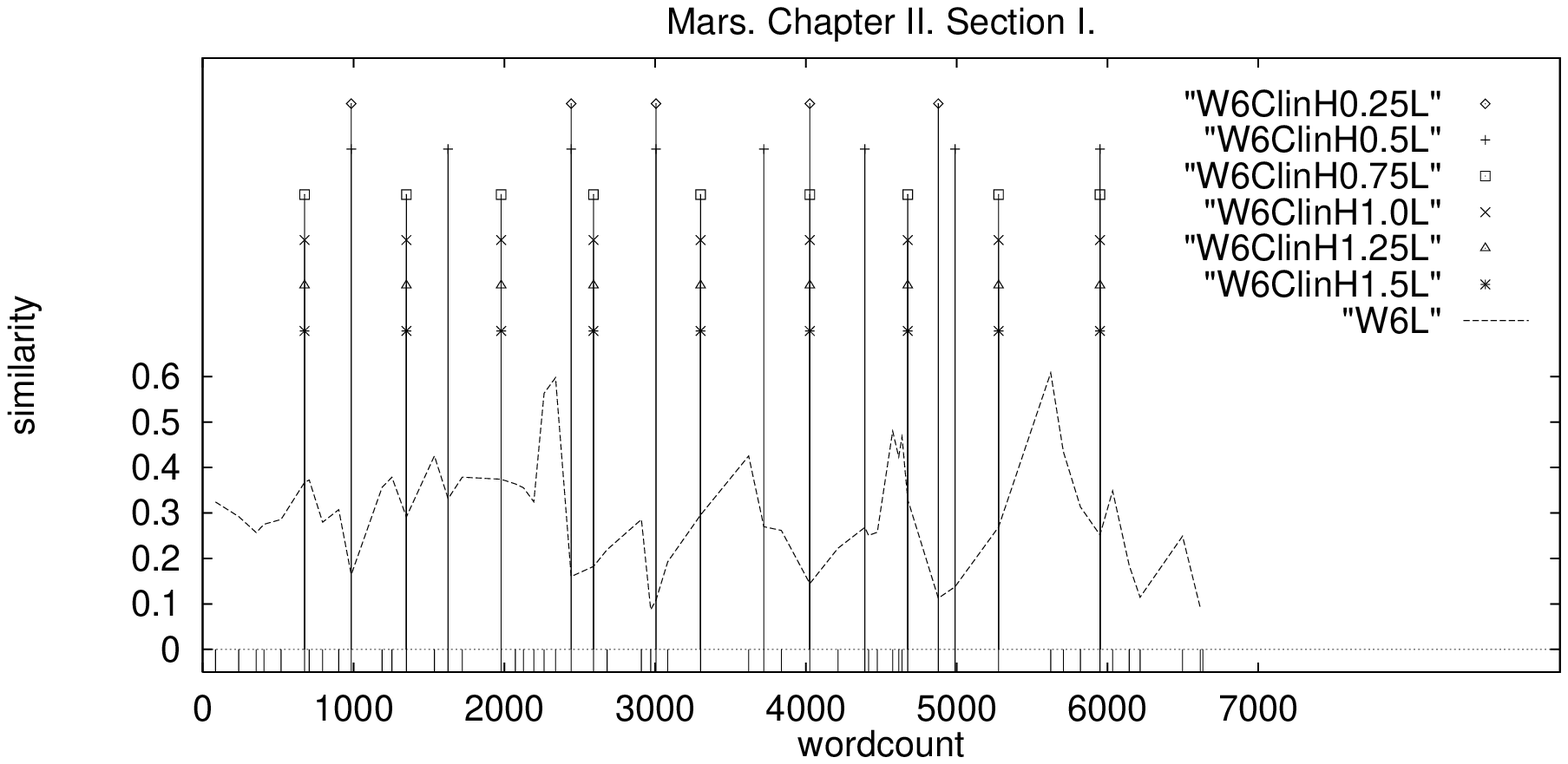}\\[1.5ex]
(a)\\[3.5ex]
\epsfxsize=\textwidth
\leavevmode
\epsffile{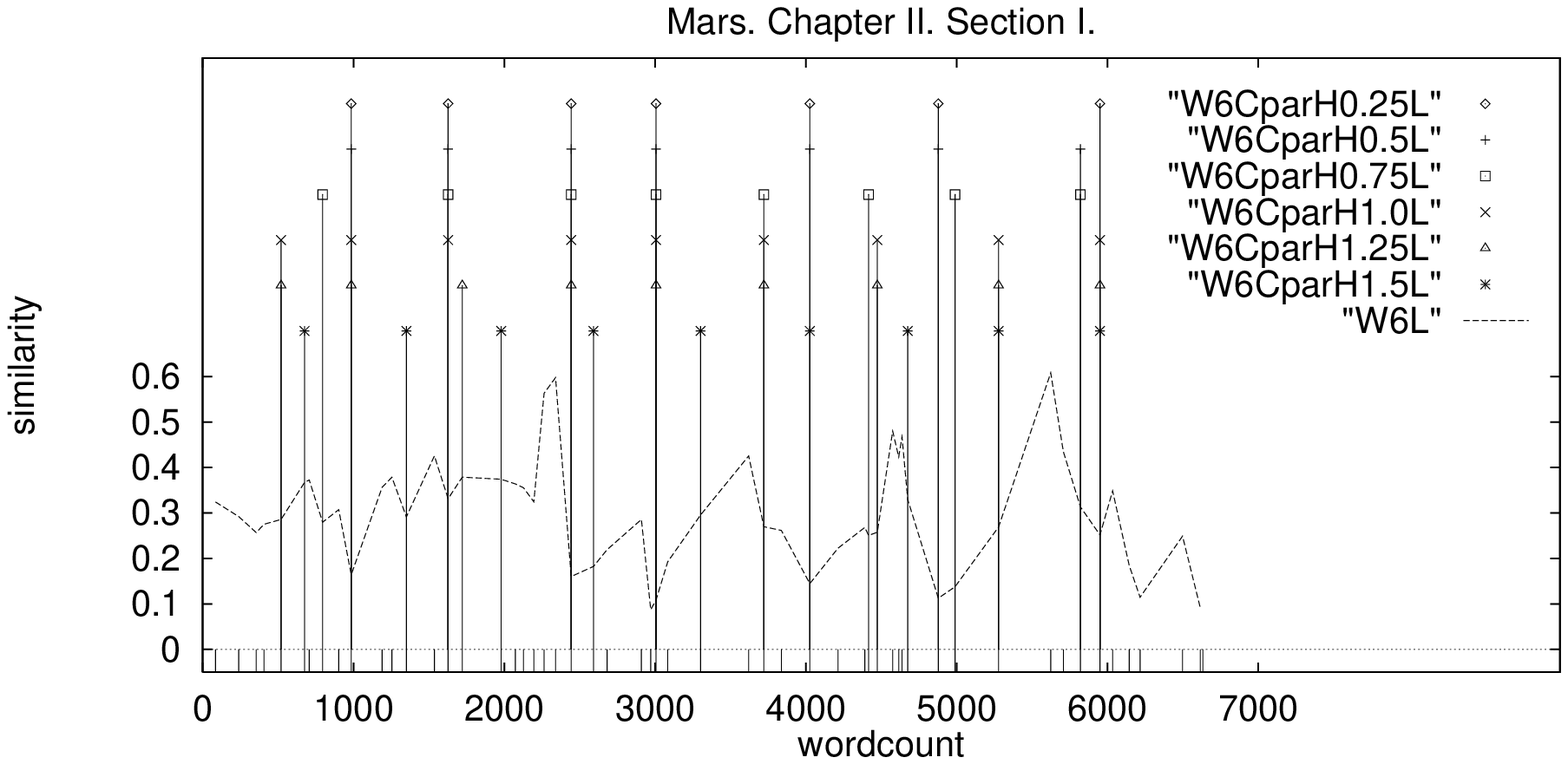}\\[1.5ex]
(b)
\end{center}
\caption{Similarity curve and detected fragment boundaries
with different cost functions. 
(a)~Linear.  (b)~Parabola.  $p$ is 600 words in both (a)~\&~(b). 
``H0.25'', etc., indicates the value of~$h$. Vertical bars indicate
fragment boundaries while short bars below horizontal axis indicate paragraph
boundaries.} 
\label{f:cost2i}
\end{figure}

\end{onecolumn}

\end{document}